\documentclass[letterpaper, 10 pt, conference]{ieeeconf}  

\IEEEoverridecommandlockouts                              
                                                          
\usepackage{cite}
\usepackage{amsmath,amssymb,amsfonts}
\usepackage{algorithmic}
\usepackage{graphicx}
\usepackage{textcomp}
\usepackage{algorithm}

\usepackage{centernot,mathtools}
\usepackage[hidelinks]{hyperref}
\usepackage[capitalize]{cleveref}
\usepackage{booktabs}
\usepackage{bm}

\usepackage{graphicx}
\usepackage{svg}

\usepackage{multicol}
\usepackage{multirow}
\usepackage{pifont}   
\usepackage{arydshln}
\newcommand{\xmark}{\ding{55}}%
\usepackage{xcolor}
\def\BibTeX{{\rm B\kern-.05em{\sc i\kern-.025em b}\kern-.08em
    T\kern-.1667em\lower.7ex\hbox{E}\kern-.125emX}}

\newcommand{\reals}{\mathbb{R}}

\newcommand{\norm}[1]{\left\lVert#1\right\rVert}
    
\begin{document}

\title{\LARGE \bf
ROSAR: An Adversarial Re-Training Framework for Robust Side-Scan Sonar Object Detection
}  


\author{Martin Aubard$^{1}$, László Antal$^{2}$, Ana Madureira$^{3}$, Luis F. Teixeira$^{4}$ and Erika Ábrahám$^{2}$
\thanks{*This project has received funding from the European Union's Horizon 2020 research and innovation programme under the Marie Skłodowska-Curie grant agreement No. 956200 (\url{https://remaro.eu}).}
\thanks{$^{1}$M. Aubard is with OceanScan Marine Systems $\&$ Technology, 4450-718 Matosinhos, Portugal,
        {\tt\small maubard@oceanscan-mst.com}.}
\thanks{$^{2}$L. Antal and E. Ábrahám are with RWTH Aachen University, 52074 Aachen, Germany,
        {\tt\small \{antal,abraham\}@cs.rwth-aachen.de}.}
\thanks{$^{3}$A. Madureira is with INESC INOV-Lab and ISRC (ISEP/P.PORTO), 4249-015 Porto, Portugal,
        {\tt\small amd@isep.ipp.pt}}
\thanks{$^{4}$L. Teixeira is with INESC TEC, Faculdade de Engenharia, Universidade do Porto, 4200-485 Porto, Portugal,
        {\tt\small luisft@fe.up.pt}}
}

\maketitle
\thispagestyle{empty}
\pagestyle{empty}
\begin{abstract}
    This paper introduces ROSAR, a novel framework enhancing the robustness of deep learning object detection models tailored for side-scan sonar (SSS) images, generated by autonomous underwater vehicles using sonar sensors. By extending our prior work on knowledge distillation (KD), this framework integrates KD with adversarial retraining to address the dual challenges of model efficiency and robustness against SSS noises. We introduce three novel, publicly available SSS datasets, capturing different sonar setups and noise conditions. We propose and formalize two SSS safety properties and utilize them to generate adversarial datasets for retraining. Through a comparative analysis of projected gradient descent (PGD) and patch-based adversarial attacks, ROSAR demonstrates significant improvements in model robustness and detection accuracy under SSS-specific conditions, enhancing the model's robustness by up to 1.85\%. ROSAR is available at \url{https://github.com/remaro-network/ROSAR-framework}.
\end{abstract}


\section{Introduction}
\label{sec:introduction}
With the growing interest in deep-sea exploration for oceanographic research \cite{b162} and energy infrastructure (e.g., gas pipelines \cite{b163}, wind turbine structures \cite{windturbine}), the development of underwater monitoring systems, particularly autonomous underwater vehicles (AUVs), has seen significant advancements over the last decade.
Due to the unique underwater environment, common sensors used in terrestrial and aerial robotics, such as cameras and LiDAR, are limited in their underwater applications. Consequently, sonar, which operates based on sound, is the most commonly used sensor underwater, overcoming limitations related to luminosity and reflection. However, despite its broad use in underwater robotics, sonar is susceptible to underwater environmental noise from other sonars, marine animals, and the deep sea. 

As the trend moves toward implementing deep learning (DL) models onboard for real-time detection and decision-making \cite{onboard-sonar}, ensuring the reliability of these models becomes crucial. Therefore, operators must trust that the DL models will consistently provide accurate detections even under such challenging conditions.
Nonetheless, ensuring this trust has been an ongoing challenge for several years due to the unpredictable underwater noise. Previous work has focused on sonar noise filtering to reduce noise in sonar images \cite{rs11040396,9931211}, which can result in information loss. Inspired by trends in generative models, current efforts focus on using generative adversarial networks (GANs) \cite{b196} to extend noisy datasets, thereby improving model robustness \cite{b198,10399359}. While useful for data enhancement, these strategies neglect to examine how the model's behavior is influenced under adverse conditions.

\begin{figure}[t]
    \centering
    \includegraphics[width=\linewidth]{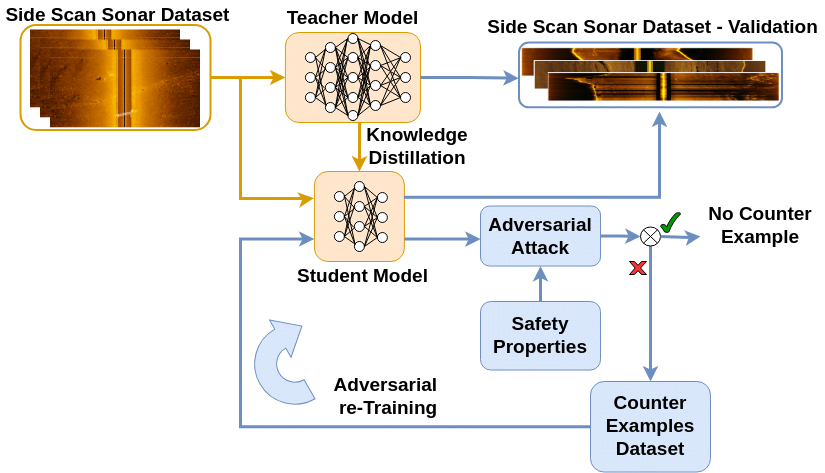} 
    \caption{The structure of the ROSAR framework. Yellow boxes show the knowledge distillation process from our previous work \cite{aubard2024knowledge}, while the blue boxes display the adversarial retraining process.}
    \label{fig:GA} 
\end{figure} 


Given the widespread use of DL, a growing research line for neural network verification (NNV) has emerged, aiming to rigorously validate DL models against specific safety and robustness criteria, contributing to significant insights into their reliability \cite{NNV}. In computer vision, NNV is commonly used for classification tasks, ensuring that the DL model consistently outputs the correct class even if a certain amount of noise is present in the input data \cite{verif-image-classification}. However, its application to more complex models like object detection remains limited, primarily due to computational constraints.

Recognizing these challenges, particularly in the context of side-scan sonar (SSS) imagery, our approach is focused on leveraging NNV (through adversarial attacks) for robustness improvement. Expanding on our previous work on knowledge distillation (KD) \cite{aubard2024knowledge} applied to the YOLOX model \cite{YOLOX}, we introduce an extended framework designed to enhance the robustness of object detection models against SSS-specific noise. This framework involves defining specific \emph{safety} or \emph{robustness properties} and retraining the model using \emph{counter-examples} (CEs) generated in cases when these properties are violated.




\cref{fig:GA} illustrates the proposed framework, where the yellow arrows and boxes highlight the KD approach from our previous work \cite{aubard2024knowledge}, while the blue arrows and boxes showcase the novel contributions introduced in this paper, including:

\begin{itemize} 
    \item Introduction of ROSAR, a novel adversarial retraining framework specifically designed to enhance the robustness of object detection models in SSS images. 
    \item Formalization of two novel safety properties tailored for underwater object detection within SSS images. 
    \item Release of three field-collected SSS datasets featuring varying noise levels alongside three adversarially generated SSS datasets. 
\end{itemize}

This paper is organized as follows: \cref{sec:related-work} reviews the state-of-the-art in adversarial attacks for both general and sonar-specific imagery, \cref{sec:methodology} details the methodology employed to implement ROSAR, \cref{sec:datasets} introduces the three new SSS datasets, \cref{sec:adversarial_attack} formalizes the SSS safety properties, \cref{sec:exp-results} presents and compares the results of the retrained models, and \cref{sec:conclusion} concludes the paper.


\section{Related Work}
\label{sec:related-work}

\textit{Object Detection} aims to accurately detect and classify objects on an image (or video). When deploying a neural network model for object detection into an embedded system, the typical trade-off is between choosing efficient (but less accurate) models and accurate (but less efficient) models. Focusing on improving the efficiency of the embedded model, our previous work \cite{aubard2024knowledge} leverages KD \cite{hinton2015distilling} to distillate the knowledge from a teacher (larger model) to a student (smaller model), achieving an improvement in the accuracy of the smaller model, while maintaining its efficiency. However, while \cite{aubard2024knowledge} focuses on knowledge distillation, it does not address the issue of model robustness. Thus, ROSAR is designed to ensure accurate output predictions, even in the presence of noises absent from the training dataset.

\textit{Adversarial attacks} on neural networks have gained significant attention in recent years, especially in safety-critical applications such as autonomous driving and industrial robotics, where the DL output prediction must be robust for safe operation. Szegedy et al. \cite{szegedy2014intriguing} first demonstrated that neural networks are vulnerable to adversarial attacks, i.e., minor perturbations to the input data can cause the model to make incorrect predictions with high confidence. Goodfellow et al. introduced the fast gradient sign method (FGSM) \cite{FGSM} to craft adversarial examples by leveraging model gradients. Unlike more straightforward methods like FGSM, which applies a single step of gradient ascent, Madry et al. \cite{PGD} developed the projected gradient descent (PGD) method. This iterative approach performs multiple iterations of small perturbations, refining adversarial attacks and enhancing the success rate of the attack.
Expanding on these concepts, Zhang et al. proposed alpha-beta-CROWN \cite{alpha-beta-crown}, a robust neural network verifier that certifies the robustness of neural networks against adversarial attacks, by combining branch-and-bound techniques with linear bounds propagation, guaranteeing tight robustness. Furthermore, as pre-check prior to complete verification, the tool uses the PGD attack as an efficient, but incomplete method for falsifying the safety of a network.


\textit{Adversarial patch attacks}, introduced by Brown et al. \cite{AdversarialPatch}, search for a specific patch that, when displayed on an image, can deceive the model in both classification and regression tasks. Unlike typical attacks, such as the PGD, which modify the entire image, the adversarial patch is a localized modification designed to cause misclassification. It does not require access to the entire image, and can be effective across various objects and scenes. 
Wu et al. \cite{RealWorldAdversarialAttacks} introduced a method for generating adversarial patches effective in both digital and real-world attacks on object detectors. This pioneering work focuses on the transferability of patch attacks across various models, including a total variation penalty to ensure patch smoothness. 
Building on these advancements, the DPatch \cite{DPATCH} method refines adversarial patch strategies by introducing targeted (predicting a specific incorrect class) and untargeted patches (causing the model to make any incorrect prediction). 
More recently, Shrestha et al. \cite{PatchYOLOv5} developed an adversarial patch specifically for the YOLOv5 model, achieving an 80\% success rate on the VisDrone dataset designed for unmanned aerial vehicle (UAV) applications. Their approach integrates total variation loss, printability loss, patch saliency loss, and patch objectiveness loss during the patch generation process, significantly enhancing the success rate of the attack against object detectors.

Surprisingly, the current object detection literature based on sonar images does not yet show significant interest in adversarial attacks, particularly in the context of adversarial patch attacks. Despite the current lack of previous works focusing on adversarial attacks for sonar images, Q. Ma et al. proposed the noise adversarial network (NAN) \cite{FasterRcnnAdversarialAttack}, which generates noise for sonar datasets and applies it to the Faster R-CNN object detection model, improving detection robustness by 8.9\% mean average precision and introduced the Lambertian adversarial sonar attack (LASA) \cite{b77} improving SSS classifier robustness.

\section{Methodology}
\label{sec:methodology}
Our proposed framework, illustrated in \cref{fig:GA}, is designed with two primary objectives: (1) leveraging KD to enhance the efficiency and accuracy of the YOLOX object detection model and (2) increasing the model's robustness against noise. While the first objective has been covered in \cite{aubard2024knowledge}, this paper centers on the second objective, which integrates the KD-enhanced model into an adversarial retraining loop. Validation is conducted on field-collected noisy SSS images, with robustness assessment using adversarial datasets generated by PGD and adversarial patch attacks.

\textit{PGD Attack.} Using the alpha-beta-CROWN tool, the PGD attack is conducted based on the safety properties defined in \cref{sec:adversarial_attack}. If the model violates these properties then the tool generates a counter-example, producing an adversarial image that triggers the violation. To characterize robustness, binary search is used to determine a noise tolerance upper bound, below which correct predictions are maintained, as detailed in \cref{PGD Attack-Adversarial Retraining}.

\textit{Adversarial Patch Attack.} This attack is implemented by focusing on the YOLOv5 model. Due to certain constraints in integrating the YOLOX model within the dedicated tool, we chose to (1) train the YOLOv5 model on the SWDD dataset and (2) apply the adversarial patch to the ground truth locations within the images from the SWDD dataset, assessing the transferability of the adversarial patch between YOLOv5 and YOLOX models.

\textit{Adversarial Retraining.} Using counter-examples generated by the two adversarial attacks (PGD and Patch), two distinct adversarial datasets, PGD-SWDD and Patch-SWDD, are generated. The adversarial retraining loop applies these datasets separately to fine-tune the original model from its last saved weights, leveraging transfer learning, and experimented with different epochs ensuring effective retraining, as presented in \cref{tab:merged-retraining}.

\textit{Robustness Validation.} After selecting the optimal retraining epoch, we again employed the PGD attack and the binary search method to validate the improvement in model robustness, comparing the original with both adversarially retrained models (PGD and Patch), as described in \cref{sec:exp-results}.


\section{Datasets}
\label{sec:datasets}

The lack of open-source sonar datasets often forces underwater robotics researchers to collect and annotate their own data, a time-consuming and expensive process that limits the ability to compare scientific results and the reproducibility of experiments \cite{electronics9111972}. Thus, to validate our proposed method, we introduce SWDD-Validation, which is composed of three novel open-source SSS datasets: SWDD-Clean, SWDD-Surface, and SWDD-Noisy, all of them extending our previously published dataset SWDD \cite{aubard2024knowledge}. The datasets are available at \url{https://zenodo.org/records/10528135}. 

\begin{figure}
  \centering
  \begin{minipage}{0.48\textwidth}
      \centering
      \textbf{SWDD-Clean}\\[0.1cm] 
      \includegraphics[width=\textwidth]{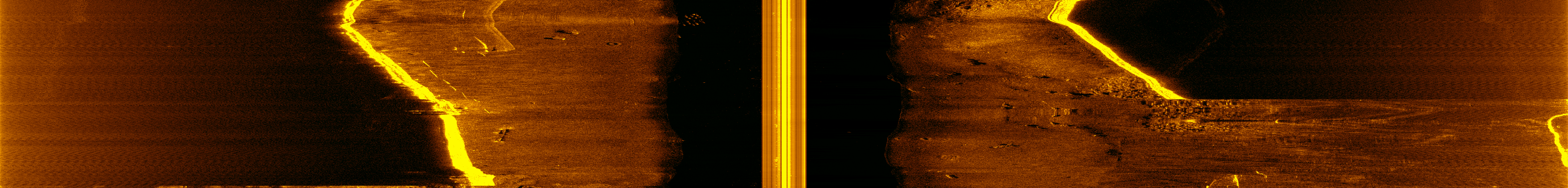}
  \end{minipage}
  \vfill
  \begin{minipage}{0.48\textwidth}
      \centering
      \textbf{SWDD-Surface}\\[0.1cm] 
      \includegraphics[width=\textwidth]{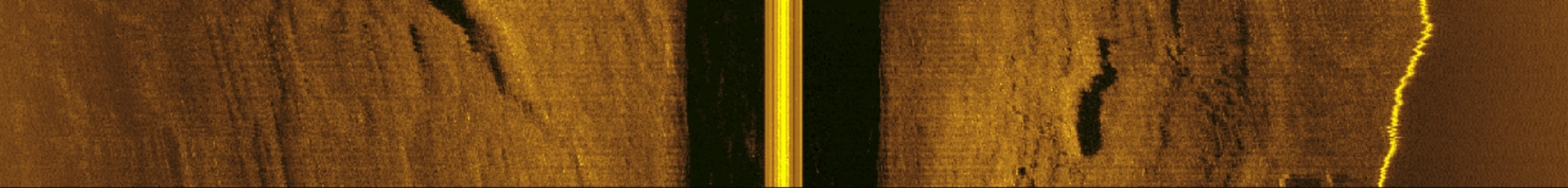}
  \end{minipage}
  \vfill
  \begin{minipage}{0.48\textwidth}
      \centering
      \textbf{SWDD-Noisy}\\[0.1cm] 
      \includegraphics[width=\textwidth]{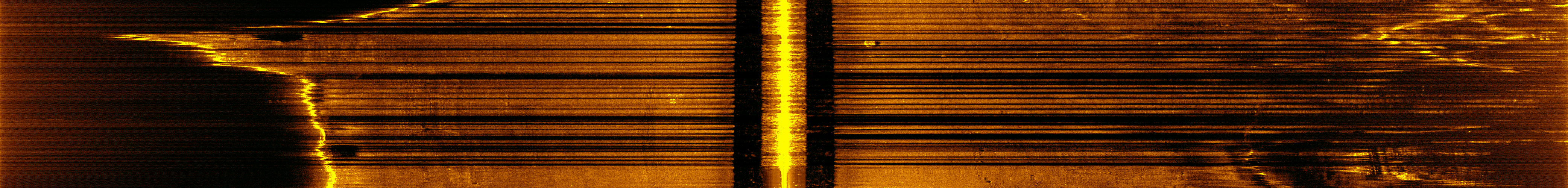}
  \end{minipage}
  \caption{Samples of SWDD-Clean, SWDD-Surface and SWDD-Noisy.}
  \label{fig:sample_SSS_dataset}
\end{figure}

In this paper, we train the original model with the SWDD dataset and validate it using the three proposed datasets, aiming to evaluate how the robustness of the model may vary under different sonar and noise conditions.
\begin{table}
  \caption{SWDD-Validation Dataset Metadata.}
  \begin{tabular}{lccccc}
    \toprule
    \textit{Dataset} & \textit{\#Image} & \textit{\#BBox} & \textit{Freq (Khz)} & \textit{Range (m)} & \textit{Resolution}        \\
    \midrule
    Clean  & 148 & 248 & 900  & 50 & $4168 \times 500$     \\
    Surface & 98 & 153 & 900  & 75 & $6552 \times 500$     \\
    Noisy  & 551 & 800 & 455 & 100 & $4168 \times 500$     \\
    \bottomrule
  \end{tabular}
  \label{tab:dataset_metadata}
\end{table}
Similarly to the SWDD dataset, the three new datasets are the results of wall inspection surveys, collected at the Porto de Leixões harbor using a Klein 3500 sonar mounted on a Light Autonomous Underwater Vehicle (LAUV) \cite{SOUSA2012268}. 
\cref{tab:dataset_metadata} describes the datasets, providing meta-data on the number of images, bounding boxes, sonar frequency, range per transducer, and the total resolution of the generated images. 
\cref{fig:sample_SSS_dataset} provides a sample image from each dataset: the SWDD-Clean dataset, which includes data from the same mission as the original SWDD dataset; the SWDD-Surface dataset, captured while the LAUV was on the surface during windy weather, featuring a non-straight wall and wave-induced variations; and the SWDD-Noisy dataset, collected under stormy conditions, where the SSS transducer intermittently exited the water, resulting in data loss represented by black lines in the images.
To clarify the features of SSS images, the yellow line in the center represents the nadir gap between the two SSS transducers, indicating areas where seafloor data is absent. Additionally, the yellow line outside the nadir gap denotes the presence of a wall.



\section{Adversarial Attack}
\label{sec:adversarial_attack}

Neural network verification (NNV) is an emerging area of formal methods that enables practitioners to mathematically prove whether certain properties hold for a given neural network.
These properties are usually defined as a set of constraints that restrict the inputs and outputs of the network. 
Given a neural network with input-output function $f : \reals^{n} \rightarrow \reals^{m}$, and a property of interest $\mathcal{P} \coloneq (\mathcal{P_{\text{in}}},  \mathcal{P_{\text{out}}})$, the goal of NNV is to check whether $\forall x \; . \; x \vdash \mathcal{P_{\text{in}}} \implies f(x) \vdash \mathcal{P_{\text{out}}}$ holds.
If it holds, then the neural network $f$ is said to satisfy $\mathcal{P}$. In the case of violation, the verification tools generally provide a counter-example, which is an input $\bar{x}$ such that $\bar{x} \vdash \mathcal{P_{\text{in}}}$ but $f(\bar{x}) \centernot \vdash \mathcal{P}_{\text{out}}$.

Various properties of interest, such as safety or robustness, can be formulated depending on the verification objective. A classical robustness property asserts that the network's output is robust against small input perturbations. More specifically, considering some $\mathcal{L}_p$ norm, an input $\bar{x} \in \reals^{n}$, and positive real constants $\varepsilon$ and $\delta$, the network $f$ is locally robust for input $\bar{x}$, iff $\forall x. \; \; \norm{x - \bar{x}}_{p} \leq \epsilon \implies \norm{f(x) - f(\bar{x})}_{p} \leq \delta$. The most commonly used norm is the $\mathcal{L}_{\infty}$ norm.

In this paper, we define and analyze two safety properties, both of which describe the robustness of our YOLOX model. Since YOLOX is an object detection model, outputting a fixed number of bounding box proposals along with objectness scores and class confidence scores for each bounding box, one needs to account for these factors when formalizing the safety properties. Accordingly, our robustness properties are designed to assess whether adversarial noise in the input images can compromise the output, leading to instances where some predicted bounding boxes are effectively fooled.


i. Our first property $\mathcal{P}_1$ expresses that the network is robust against random $\mathcal{L}_{\infty}$ noise in the input. This type of noise simulates the random perturbations that can be present in side-scan sonar images across the whole waterfall image. Our constraints allow noise in each pixel and each channel by a portion of $0<\varepsilon<1$. 
The property is violated in case there is a noisy input, within the $\varepsilon$ perturbation bound, for which either the predicted bounding box objectness score falls below the objectness threshold $\xi^{\text{obj}}$ or the predicted class for the bounding box changes.
Formally, for a 3D input image $\bar{x}$ of size $h {\times} w {\times} c$ and a maximum perturbation bound $\varepsilon$, attacking the bounding box $b$, with objectness score $y_{b}^{\text{obj}}$, having $N$ classification confidence scores and $y_{b}^{\text{class}_p}$ being the confidence score of the correct class, the robustness property is defined as follows: 
\begin{equation*}
    {\small
    \label{equ:P1}
    \begin{aligned}
        \mathcal{P}_1 \coloneq  & \bigwedge_{i, j, k = 1, 1, 1}^{h, w, c}  (1 - \epsilon) \cdot \bar{x}_{i, j, k} \leq x_{i, j, k} \leq (1 + \epsilon) \cdot \bar{x}_{i, j, k} \\
        & \implies \left(y_{b}^{\text{obj}} \geq \xi^{\text{obj}} \land \left(\bigwedge_{l=1, l \neq p}^{N} y_{b}^{\text{class}_p} > y_{b}^{\text{class}_l}\right) \right) . 
    \end{aligned}
    }%
\end{equation*}
ii. Our second property $\mathcal{P}_2$ expresses that the network is robust against dark horizontal lines in the input image, mimicking the noise that is present in the SWDD-Noisy dataset. 
To verify the model performance against this black line phenomenon, we formalize the robustness property $\mathcal{P}_2$ for a randomly generated line configuration
$L \subseteq\{1,\ldots,h\}$ and some $0<\varepsilon<1$ as follows:
\begin{equation*}
    {\small
    \label{equ:P2}
    \begin{aligned}
        \mathcal{P}_2 \coloneq  & \hspace*{-0.15cm} \bigwedge_{j=1,k=1}^{w,c} \Big[\Big(
        \big(\bigwedge_{i \in L} \epsilon \cdot \bar{x}_{i,j,k} \leq x_{i,j,k} \leq \bar{x}_{i,j,k}\big) \wedge\\
        & \big(\hspace*{-1.5ex}\bigwedge_{i \in \{1,\ldots,h\}\setminus L} \hspace*{-3ex} x_{i,j,k} = \bar{x}_{i,j,k}\big)
        \Big) \implies \\
        & \Big(y_{b}^{\text{obj}} \geq \xi^{\text{obj}} \land \big(\bigwedge_{l=1, l \neq p}^{N} y_{b}^{\text{class}_p} > y_{b}^{\text{class}_l}\big) \Big) \Big].
    \end{aligned}
    }%
\end{equation*}

As an alternative to NNV, adversarial attacks offer an incomplete but often more efficient way of falsifying the robustness of neural networks. An adversarial attack tries to find the input $\bar{x}$, which violates the robustness property from above, resulting in unexpected output, i.e., \textit{fooling} the network. Since the formal verification of a neural network is an NP-complete problem, analyzing real-world-sized networks, such as YOLOX and other object detection models, is a challenging and in most cases infeasible task (considering limited amount of resources). Thus, in this paper we only provide insights into the robustness of the networks by assessing the success rate of different adversarial attack methods against them. Using adversarial attacks, we can easily show \emph{unsafety} (i.e. the lack of adversarial robustness), which is the case in most instances. However, the result of this analysis is not a formal guarantee due to the incompleteness of these methods.

\section{Experimental Evaluation}
\label{sec:exp-results}
As outlined in \cref{sec:methodology}, the adversarially retrained models are validated using two approaches: (1) the SWDD-Validation datasets to assess the improvement of the retrained model compared to the baseline results (\cref{tab:merged-retraining}), and (2) the PGD attack embedded in binary search to compute the robustness bounds considering the \(\mathcal{P}_1\) and \(\mathcal{P}_2\) safety properties. Both validation processes are conducted using the KD-Nano-L-ViT model \cite{aubard2024knowledge}, resulting from the KD of the YOLOX-ViT-L model into the YOLOX-Nano model.

\subsection{Adversarial Dataset Generation}

Adversarial retraining begins with creating adversarial datasets, which are subsequently integrated into the retraining loop to enhance model robustness. This section details the generation of these adversarial datasets and evaluates the retrained model under both PGD and patch attack scenarios.

\subsection*{PGD - Adversarial Dataset}
\label{PGD Attack-Adversarial Retraining}

Our study uses the alpha-beta-CROWN tool to assess whether the PGD attack can produce a counter-example, signifying a violated safety property. Due to current computational constraints, we cannot verify the safety property to ensure complete compliance. However, based on extensive testing, we consider the safety property satisfied if the PGD attack does not produce a counter-example within two minutes. We employ a binary search method to approximate the noise threshold at which the model fails.

\begin{algorithm}
  \scriptsize
  \caption{Binary search to find the adversarial bound}
  \label{alg:binary_search_ce}
  \begin{algorithmic}[1]
      \STATE \textbf{Input:} $\mathcal{P}$, $\mathit{dataset}$, $L$, $U$, $\mathit{max\_iter}$, $\mathit{time\_limit}$ 
      \STATE \textbf{Output:} Threshold value of each instance where safety property fails
      \FOR{each $\mathit{img}$ \textbf{in} $\mathit{dataset}$}
          \FOR{each $\mathit{bbox}$ \textbf{in} $\mathit{inference}(\mathit{img})$}
              \STATE $\mathit{low} \gets L$, $\mathit{high} \gets U$
              \FOR{$i$ \textbf{from} $1$ \textbf{to} $\mathit{max\_iter}$}
                  \STATE $\mathit{mid} \gets \frac{\mathit{low} + \mathit{high}}{2}$
                  \STATE $\mathit{found\_CE}, \mathit{CE} \gets \mathit{eval\_prop}(\mathcal{P}, \mathit{img}, \mathit{bbox}, \mathit{mid}, \mathit{time\_limit})$
                  \IF{$\mathit{found\_CE}$} 
                    \STATE $\mathit{high} \gets \mathit{mid}$
                    \STATE $\mathit{save\_image}(\mathit{CE})$
                  \ELSE
                    \STATE $\mathit{low} \gets \mathit{mid}$
                  \ENDIF
                  \STATE Save and report threshold value $high$ of the verification instance
              \ENDFOR
          \ENDFOR
      \ENDFOR
  \end{algorithmic}
  \end{algorithm}

The binary search, outlined in \cref{alg:binary_search_ce} is applied to both safety properties ($\mathcal{P}_1$ and $\mathcal{P}_2$), where the function $\mathit{eval\_prop}$ takes as input the safety property. The input parameters differ depending on the property: for \(\mathcal{P}_1\), the lower and upper bounds are set to 0.0 and 0.08, respectively, with a maximum of 5 iterations; for \(\mathcal{P}_2\), the bounds are 0.60 and 1.0, with also up to 5 iterations. The algorithm iterates over all detected bounding boxes for each input image, initializing the search bounds. The midpoint ($mid$) is calculated by bisecting the interval in each iteration. The property check, performed by the $\mathit{eval\_prop}$ function, is conducted for the perturbation bound $mid$. If a counter-example is found within the specified time limit, the upper bound is adjusted downward, and the search range is halved. If no counter-example is found, the lower bound is adjusted upward accordingly. This iterative process continues for the designated number of iterations, with the final threshold bound saved as the average of the maximal perturbation bound where the property holds and the minimal perturbation bound where the property fails. The counter-examples found during each iteration are saved in the allocated adversarial dataset, resulting in two separate datasets -- one for \(\mathcal{P}_1\) and one for \(\mathcal{P}_2\), named P1-SWDD (1017 images) and P2-SWDD (1462 images). A sample image from each dataset are displayed on \cref{fig:Adversarial-SWDD} (left and middle).

\subsection*{Patch - Adversarial Dataset}

\begin{figure}[t]
  \centering
  \begin{tabular}{c@{\hspace{0.6em}}c@{\hspace{0.6em}}c}
   P1-SWDD & P2-SWDD & Patch-SWDD\\ 
  
  \includegraphics[width=0.15\textwidth]{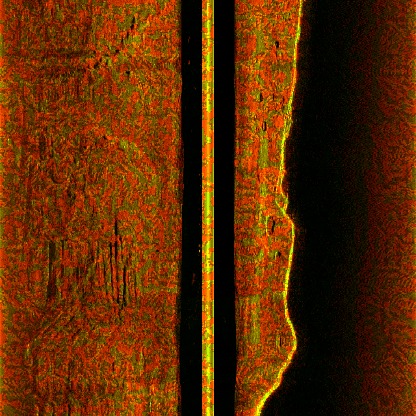} & \includegraphics[width=0.15\textwidth]{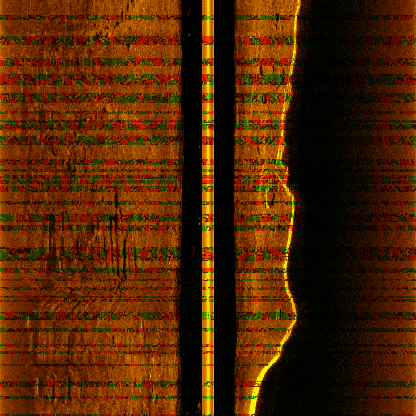} & \includegraphics[width=0.15\textwidth]{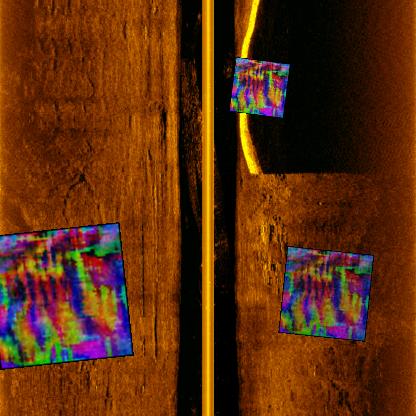}\\
  \end{tabular}
  
  \caption{Sample images of the three adversarial datasets.}
  \label{fig:Adversarial-SWDD}
\end{figure}

Due to some limitations of integrating YOLOX into the patch generation framework, for this experiment, we opted to use the YOLOv5 model for adversarial patch dataset generation and subsequently apply this dataset in the adversarial retraining loop using the KD-Nano-L-ViT model.
The adversarial patch attack on the YOLOv5 model requires initial training with the SWDD dataset. To align with the size of the YOLOX model used in this study, we select the YOLOv5-nano model and train it for 300 epochs. The resulting model weights are then incorporated into the adversarial patch framework, as explained in \cite{PatchYOLOv5}. 
By applying this method, ROSAR generates the adversarial Patch-SWDD dataset, which consists of 151 images. The adversarial dataset comprises the SWDD dataset with the adversarial patch in the dataset ground truth location for every bounding box, corresponding for the classes \textit{wall} and \textit{noWall}, as represented in the last image of \cref{fig:Adversarial-SWDD}

\subsection{Adversarial Retraining}

\begin{table}[t]
  \caption{Retraining evaluation of KD-Nano-L-ViT with Adversarial Patch, P1-SWDD, and P2-SWDD Images.}
  \begin{tabular}{lc@{\hspace{0.6em}}|@{\hspace{0.6em}}c@{\hspace{0.6em}}c@{\hspace{0.6em}}c@{\hspace{0.2em}}|@{\hspace{0.6em}}c@{\hspace{0.6em}}c@{\hspace{0.6em}}c@{\hspace{0.6em}}|@{\hspace{0.6em}}c@{\hspace{0.6em}}c@{\hspace{0.6em}}c}
    \toprule
    & \multicolumn{1}{c|}{} & \multicolumn{3}{c|}{\textbf{Patch-SWDD}} & \multicolumn{3}{c|}{\textbf{P1-SWDD}} & \multicolumn{3}{c}{\textbf{P2-SWDD}} \\
    \cmidrule(lr){3-5} \cmidrule(lr){6-8} \cmidrule(lr){9-11}
    \textit{Val.} & \textit{epoch} & \textit{\%TP} & \textit{FP} & \textit{AP} & \textit{\%TP} & \textit{FP} & \textit{AP} & \textit{\%TP} & \textit{FP} & \textit{AP} \\
    \midrule
    \multirow{5}{*}{Clean} & \xmark & 82 & 76 & 0.64 & - & - & - & - & - & - \\
    & 5 & \textbf{75} & 59 & 0.64 & 50 & 14 & 0.67 & 73 & 122 & 0.51 \\
    & 10 & 44 & 9  & \textbf{0.66} & 51 & \textbf{11} & \textbf{0.69} & 49 & 42 & 0.55 \\
    & 15 & 39 & 4  & \textbf{0.66} & 54 & 15 & 0.68 & 53 & \textbf{24} & 0.64 \\
    & 20 & 28 & \textbf{3} & 0.61 & \textbf{57} & 24 & 0.67 & \textbf{74} & 46 & \textbf{0.68} \\
    \midrule
    \multirow{5}{*}{Surface} & \xmark & 59 & 28 & 0.60 & - & - & - & - & - & - \\
    & 5 & \textbf{51} & \textbf{0} & 0.64 & 29 & \textbf{0} & 0.64 & \textbf{57} & 35 & 0.56 \\
    & 10 & 25 & \textbf{0} & 0.62 & 33 & \textbf{0} & 0.66 & 28 & \textbf{0} & 0.67 \\
    & 15 & 30 & \textbf{0} & 0.65 & 37 & \textbf{0}  & \textbf{0.68} & 36 & \textbf{0} & 0.67 \\
    & 20 & 38 & \textbf{0} & \textbf{0.69} & \textbf{43} & 33 & 0.47 & 44 & \textbf{0} & \textbf{0.72} \\
    \midrule
    \multirow{5}{*}{Noisy} & \xmark & 74 & 143 & 0.69 & - & - & - & - & - & - \\
    & 5 & \textbf{72} & 165 & 0.66 & 48 & 27 & 0.68 & \textbf{77} & 207 & 0.65 \\
    & 10 & 39 & \textbf{10} & \textbf{0.67} & 41 & 31 & 0.64 & 42 & 138 & 0.50 \\
    & 15 & 42 & 18  & \textbf{0.67} & 53 & 32 & \textbf{0.71} & 56 & \textbf{46} & \textbf{0.70} \\
    & 20 & 44 & 28  & 0.66 & \textbf{58} & 58 & 0.70 & 66 & 130 & 0.66 \\
    \bottomrule
  \end{tabular}
  \label{tab:merged-retraining}
\end{table}

We applied adversarial retraining with the three adversarial datasets to fine-tune the KD-Nano-L-ViT model, initially trained on the SWDD dataset for 300 epochs. To enhance the model's robustness, the retraining process leverages transfer learning with the P1-SWDD, P2-SWDD, and Patch-SWDD datasets. Since the retraining process focuses on adversarial retraining rather than initial training, we aim to make the model retain the knowledge acquired during the initial training. Consequently, the retraining is conducted by comparing the performance across four different epochs: 5, 10, 15, and 20 epochs. The results of the adversarial retraining are illustrated in \cref{tab:merged-retraining}, where \textit{Val.} indicates the validation dataset, \textit{epoch} specifies the number of epochs used for retraining, \textit{$\%$TP} is the percentage of true positive bounding boxes, \textit{FP} is the number of false positive bounding boxes, and \textit{AP} is the average precision. Furthermore, the first row of each validation dataset represents the metrics for the original KD-Nano-L-ViT model trained with the SWDD dataset (repetitions marked by "-").

The results demonstrate how adversarial retraining -- employing both adversarial patch and PGD methods -- enhanced the model's performance across various metrics. Notably, there is an improvement in the model's performance on all three validation datasets (SWDD-Clean, SWDD-Surface, and SWDD-Noisy). While the retrained models exhibit a reduction in \%TP, they also show a marked decrease in FP, indicating a reduction in overfitting, suggesting that the retrained models offer more reliable detections than the original.
The patch retraining has a lower \%TP than the two other models, where the P2-SWDD has the highest \%TP. Thus, as an inference comparison with the SWDD-Validation dataset, the two PGD retraining datasets have higher \%TP, whereas the patch retraining dataset has the lowest FP. Based on the results from \cref{tab:merged-retraining}, for robustness validation we have chosen the three models retrained with 15 epochs.  

\begin{figure*}[t]
  \centering
  \includegraphics[width=\linewidth, trim={3.25cm 0.40cm 3.00cm 1.20cm},clip]{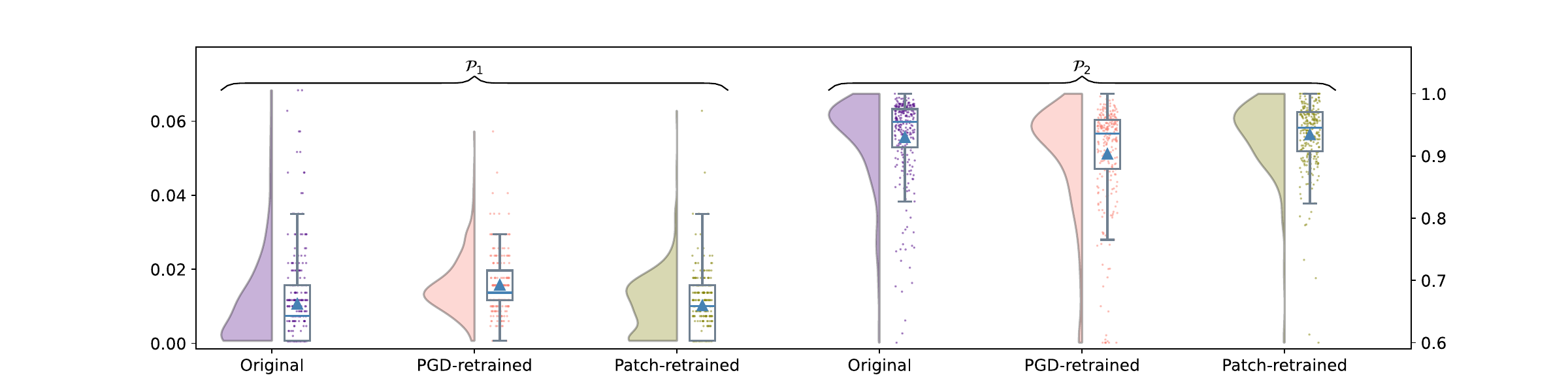}
  \caption{Robustness validation for $\mathcal{P}_1$ and $\mathcal{P}_2$.} 
  \vspace*{-2ex}
  \label{fig:robust-val-P1-P2} 
\end{figure*}

\subsection{Robustness Validation}
\label{Robustness-Validation}

The robustness validation process evaluates whether the retrained models have enhanced performance compared to the original KD-Nano-L-ViT model concerning the properties $\mathcal{P}_1$ and $\mathcal{P}_2$. This process uses the PGD attack with the aim to fool the model considering the two safety properties. Similarly to the approach in \cref{PGD Attack-Adversarial Retraining}, where counter-examples were generated, for each successful attack, we establish the threshold noise level at which the model was fooled. This phase focuses on determining the robustness boundary value for the adversarially retrained model using the binary search method in \cref{alg:binary_search_ce}. Respectively to the used property, the model retrained on the P1-SWDD dataset is compared to the original model under property $\mathcal{P}_1$, and the model retrained on the P2-SWDD is compared with the original model under property $\mathcal{P}_2$. The model retrained on the Patch-SWDD is compared using both safety properties due to the patch attack disregarding the safety properties.


\begin{table}[b]
  \centering
  \caption{Mean and median values for $\mathcal{P}_1$ and $\mathcal{P}_2$.}
  \label{tab:adv-result}
  \begin{tabular}{llccc}
  \toprule
  \textbf{Robust.} & \textbf{Metric} & \textbf{Original} & \textbf{PGD-SWDD} & \textbf{Patch-SWDD} \\
  \midrule
  \multirow{2}{*}{${\mathcal{P}_1}$} & Mean   & 0.0107 & \textbf{0.0157} & 0.0098 \\
  & Median & 0.0073 & \textbf{0.0137} & 0.0100 \\
  \hdashline
  \multirow{2}{*}{${\mathcal{P}_2}$} & Mean   & 0.9302 & \textbf{0.9026} & 0.9317 \\
  & Median & 0.9544 & \textbf{0.9359} & 0.9457 \\
  \bottomrule
  \end{tabular}
\end{table}

\cref{fig:robust-val-P1-P2} provides raincloud plots for evaluating the robustness of the original, PGD-SWDD and Patch-SWDD models, under $\mathcal{P}_1$ (on the left) and $\mathcal{P}_2$ (on the right). The violin plots show the distribution of robustness boundary values, highlighting the spread and density of the data, which helps in understanding how frequently certain robustness levels occur. The box plots to the right of the violin plots offer a clear summary of the data, indicating the mean (blue triangle), median (blue line), first and third (top and bottom lines of the box) quartiles, making it easier to compare the central robustness tendencies between models. The mean and median values are also displayed in \cref{tab:adv-result}. Lastly, the raw data is illustrated as the strip plot underlaid of each box.

$\bm{\mathcal{P}_1}$. For property $\mathcal{P}_1$, the higher the $\varepsilon$ value is, the higher the maximal noise in the data. The violin plots for the \(\mathcal{P}_1\) robustness validation indicate that while the original model exhibited the lowest median values, suggesting lower robustness overall, the retrained models showed improvement. However, despite the increase in robustness, the Patch-SWDD model displayed slightly lower mean robustness than the original model (-0.009), suggesting that the Patch-SWDD model has greater robustness stability, as its robustness is more consistent across different instances. In contrast, the original model, although capable of higher robustness in some cases, lacks this stability. Regarding the PGD-SWDD model, it demonstrates improvements in both the mean (+0.005) and median (+0.0064) metrics, reflecting enhanced robustness and stability under the \(\mathcal{P}_1\) safety property.
$\bm{\mathcal{P}_2}$. Based on the \(\mathcal{P}_2\) property from \cref{equ:P2}, the lower the $\varepsilon$ value is, the more noise is allowed in the data. The violin plots for \(\mathcal{P}_2\) show that the Patch-SWDD model, despite having slightly higher mean value (+0.015), has a median value that is lower than the original model (-0.0087), indicating that it is generally more robust across most instances. However, while displaying higher robustness in some cases, the original model shows less consistent performance overall. In contrast, the PGD-SWDD model exhibits further improved robustness, with reductions in both mean (-0.0281) and median (-0.0185) values, confirming that it offers a more stable and robust response to adversarial noise under the \(\mathcal{P}_2\) safety property.

In comparing the results from the \(\mathcal{P}_1\) and \(\mathcal{P}_2\) robustness validations, a clear pattern emerges that highlights the strengths and trade-offs of the retrained models. The PGD-SWDD model significantly improved mean and median robustness values, indicating that adversarial retraining effectively enhanced the model's ability to resist noise. Although the Patch-SWDD model showed a slightly lower mean robustness than the original model, it still provided greater stability, as evidenced by its consistent robustness across different instances.


\section{Conclusion}
\label{sec:conclusion}

This paper presented ROSAR, a novel framework to enhance the robustness and efficiency of DL object detection models tailored explicitly for SSS images. The framework leverages KD for embedded systems, previously validated in our earlier work, while focusing on improving model robustness through adversarial retraining. We addressed the challenges of SSS-specific noise and limited data availability by introducing three distinct SSS datasets and generating adversarial datasets using PGD and patch attacks. Our extensive experiments demonstrate that adversarial retraining improves detection accuracy and robustness under SSS conditions and that model retraining with PGD attack returns better model robustness. 
While the Patch-SWDD dataset slightly reduced mean robustness compared to the original model, it significantly improved detection metrics and provided greater stability, ensuring consistent robustness across various instances. Given the computational constraints, our current methodology focused on fooling the bounding box candidate with the highest confidence value. Future work should expand this approach to consider multiple candidates simultaneously, thereby providing a more comprehensive robustness assessment. Furthermore, ROSAR can be adapted to address additional safety properties, such as interference caused by data transmission during SSS data collection. This framework is not limited to SSS application but can be applied to any vision-based applications where safety properties can be mathematically formalized. This research lays a solid foundation for advancing the use of DL models in underwater robotics, particularly in challenging SSS environments.





\bibliographystyle{ieeetr}
\bibliography{biblio}

\end{document}